\documentclass[final,5p,times,twocolumn]{elsarticle}
\usepackage{etoolbox}
\AtBeginEnvironment{thebibliography}{%
  \footnotesize                 
  \setlength{\bibsep}{0pt}
  \setlength{\itemsep}{0pt}  
  \setlength{\parsep}{0pt}
  \setlength{\parskip}{0pt}
}
\usepackage{hyperref}

\usepackage{amssymb}
\usepackage{amsmath}

\begin{document}

\begin{frontmatter}



\title{AI-Generated Content in Cross-Domain Applications: Research Trends, Challenges and Propositions}

\author[label1]{Jianxin Li}
\ead{jianxin.li@ecu.edu.au}

\author[label1]{Liang Qu\corref{cor1}}
\ead{l.qu@ecu.edu.au}
\cortext[cor1]{Corresponding author.}

\author[label2]{Taotao Cai}
\ead{Taotao.Cai@unisq.edu.au}

\author[label3]{Zhixue Zhao}
\ead{zhixue.zhao@sheffield.ac.uk}

\author[label4]{Nur Al Hasan Haldar}
\ead{Nur.Haldar@curtin.edu.au}

\author[label4]{Aneesh Krishna}
\ead{A.Krishna@curtin.edu.au}

\author[label10]{Xiangjie Kong}
\ead{xjkong@zjut.edu.cn}

\author[label1]{Flavio Romero Macau}
\ead{f.macau@ecu.edu.au}

\author[label5]{Tanmoy Chakraborty}
\ead{tanchak@iitd.ac.in}

\author[label5]{Aniket Deroy}
\ead{roydanik18@gmail.com}

\author[label6]{Binshan Lin}
\ead{binshan.lin@lsus.edu}

\author[label7]{Karen Blackmore}
\ead{karen.blackmore@newcastle.edu.au}

\author[label7]{Nasimul Noman}
\ead{nasimul.noman@newcastle.edu.au}

\author[label8]{Jingxian Cheng}
\ead{chengjx3@chd.edu.cn}

\author[label8]{Ningning Cui}
\ead{csnncui@chd.edu.cn}

\author[label9]{Jianliang Xu}
\ead{xujl@comp.hkbu.edu.hk}

\affiliation[label1]{organization={School of Business and Law, Edith Cowan University},
            city={Perth},
            postcode={6027}, 
            state={Western Australia},
            country={Australia}}

\affiliation[label2]{organization={School of Mathematics, Physics and Computing, University of Southern Queensland},
            city={Toowoomba},
            postcode={4350}, 
            state={Queensland},
            country={Australia}}

\affiliation[label3]{organization={School of Computer Science, The University of Sheffield},
            city={Sheffield},
            postcode={S10 2TN}, 
            state={South Yorkshire},
            country={United Kingdom}}

\affiliation[label4]{organization={School of Electrical Engineering, Computing and Mathematical Sciences, Curtin University},
            city={Perth},
            postcode={6102}, 
            state={Western Australia},
            country={Australia}}

\affiliation[label10]{organization={College of Computer Science and Technology, Zhejiang University of Technology},
            city={Hang Zhou},
            country={China}}
            
\affiliation[label5]{organization={Department of Electrical Engineering, Indian Institute of Technology Delhi},
            city={New Delhi},
            postcode={110016}, 
            state={Delhi},
            country={India}}

\affiliation[label6]{organization={College of Business, Louisiana State University Shreveport},
            city={Shreveport},
            postcode={71115}, 
            state={Louisiana},
            country={United States}}

\affiliation[label7]{organization={School of Information and Physical Sciences, The University of Newcastle},
            city={Callaghan},
            postcode={2308}, 
            state={New South Wales},
            country={Australia}}

\affiliation[label8]{organization={School of Data Science and Artificial Intelligence, Chang'an University},
            city={Xian},
            postcode={710018}, 
            state={Shaanxi},
            country={China}}

\affiliation[label9]{organization={Department of Computer Science, Hong Kong Baptist University},
            city={Hong Kong SAR},
            country={China}}
\begin{abstract}
Artificial Intelligence Generated Content (AIGC) has rapidly emerged with the capability to generate different forms of content, including text, images, videos, and other modalities, which can achieve a quality similar to content created by humans.
As a result, AIGC is now widely applied across various domains such as digital marketing, education, and public health, and has shown promising results by enhancing content creation efficiency and improving information delivery.
However, there are few studies that explore the latest progress and emerging challenges of AIGC across different domains.
To bridge this gap, this paper brings together 16 scholars from multiple disciplines to provide a cross-domain perspective on the trends and challenges of AIGC. 
Specifically, the contributions of this paper are threefold:
(1) It first provides a broader overview of AIGC, spanning the training techniques of Generative AI, detection methods, and both the spread and use of AI-generated content across digital platforms.
(2) It then introduces the societal impacts of AIGC across diverse domains, along with a review of existing methods employed in these contexts.
(3) Finally, it discusses the key technical challenges and presents research propositions to guide future work.
Through these contributions, this vision paper seeks to offer readers a cross-domain perspective on AIGC, providing insights into its current research trends, ongoing challenges, and future directions.
\end{abstract}




\end{frontmatter}



\section{Introduction 
} 
Artificial Intelligence Generated Content (AIGC) represents a transformative trend in artificial intelligence, wherein models autonomously generate content such as text, images, audio, and video with a level of fluency and coherence approaching human ability. This technological shift is underpinned by the broader field of Generative Artificial Intelligence (Generative AI), which encompasses a variety of algorithms designed to create new data samples from learned distributions. The roots of Generative AI can be traced to early probabilistic models, including Hidden Markov Models (HMMs)~\cite{knill1997hidden}, Gaussian Mixture Models (GMMs)~\cite{reynolds2015gaussian}, and N-gram models~\cite{bengio2003neural}, which laid the groundwork for machine-generated language and speech. As neural network research advanced, architectures such as Recurrent Neural Networks (RNNs)~\cite{mikolov2010recurrent}, Long Short-Term Memory (LSTM) units~\cite{Graves_2012}, and Gated Recurrent Units (GRUs)~\cite{dey2017gate} enabled improved modeling of sequential data. In the domain of image generation, methods such as autoencoders and Variational Autoencoders (VAEs)~\cite{kingma2013auto} introduced latent variable modeling, while Generative Adversarial Networks (GANs)~\cite{goodfellow2020generative,ho2021dp,ho2019generative,qu2021imgagn} revolutionized the field with their adversarial training framework. More recently, diffusion models~\cite{song2019generative} have provided high-fidelity generation through iterative denoising processes. However, the most significant paradigm shift occurred with the emergence of transformer-based architectures~\cite{vaswani2017attention}, which unified generative modeling across modalities and enabled models to scale in size and capability.

Among these advancements, Large Language Models (LLMs)~\cite{zhao2023survey} have become the most influential and widely adopted form of generative models. Built upon the Transformer architecture, LLMs such as GPT-3~\cite{brown2020language}, LLaMA~\cite{touvron2023llama}, and ChatGPT are trained on massive corpora to predict the next token in a sequence, thereby implicitly learning syntax, semantics, and factual knowledge. Their training typically follows a pre-training and fine-tuning paradigm: first acquiring general language understanding through self-supervised learning, then adapting to specific tasks via supervised fine-tuning or prompt-based conditioning. As model scale increases--both in terms of parameters and training data--LLMs exhibit emergent capabilities such as in-context learning, logical reasoning, and summarization, all without explicit retraining. These emergent properties have fueled rapid advancements in human-computer interaction, enabling models to perform a diverse array of tasks through natural language interfaces.

LLMs have swiftly permeated various sectors including digital marketing, healthcare, education, legal services, and software development. For instance, in digital marketing, LLMs support real-time content generation, personalized advertisement design, and intelligent customer engagement. In scientific and technical domains, they assist with literature reviews, data analysis, and even software coding. Their adaptability stems from prompt-based learning, which allows them to handle downstream tasks under zero-shot, few-shot, or fine-tuned regimes with minimal labeled data. The ability to generate high-quality language and understand user intent has unlocked new modes of human-AI collaboration, while simultaneously raising important concerns around trustworthiness, disinformation, intellectual property, and societal impact.

The rest of this article is organized as follows. In Section \ref{sec:overview}, we provide an overview of large language models, including their training process, pre-training strategies, and prompt-based usage patterns. Section \ref{sec:detection} explores the generation and detection of AI-generated content. Section \ref{sec:spread} discusses the spread and use of such content across digital platforms. Sections \ref{sec:publictrust} through \ref{sec:education} examine the societal impacts of AI-generated content, covering its influence on public trust, digital marketing, public health, organizational behavior, and education. Section \ref{sec:datasovereignty} addresses issues of data sovereignty and security risks associated with generative AI. Finally, Section \ref{sec:conclusion} concludes the article. 

\section{Overview of Generative AI Framework 
}\label{sec:overview}
\subsection{Framework Overview of Generative AI}
\begin{figure*}
    \centering
    \includegraphics[width=1\linewidth]{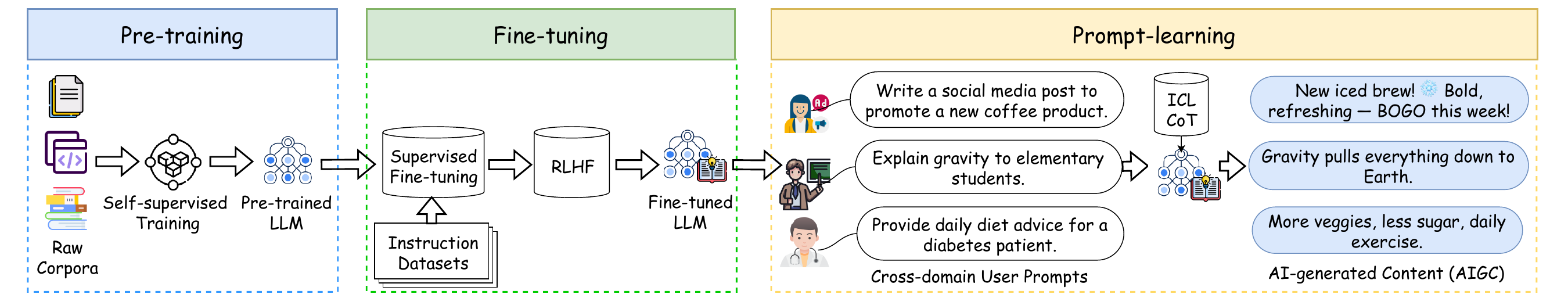}
    \caption{Illustration of the generative AI training and utilization pipeline.}
    \label{fig:aigcoverview}
\end{figure*}
The development of generative AI systems based on large language models (LLMs) generally follows a two-stage framework: training and utilization, as illustrated in Figure \ref{fig:aigcoverview} . In the training stage, models first undergo pre-training using massive unlabeled corpora to acquire fundamental language abilities and general world knowledge in a task-agnostic manner. This is followed, when necessary, by fine-tuning on domain-specific or task-specific data to improve performance in targeted applications. The utilization stage leverages prompt-based learning, where the pre-trained model is guided through natural language instructions or examples to perform downstream tasks without modifying model parameters. This paradigm enables efficient adaptation to new tasks, especially in zero-shot or few-shot settings, and forms the operational foundation of generative AI across a variety of domains. Together, these stages reflect a scalable and modular pipeline that balances general-purpose learning with task-specific alignment.

\subsection{Generative AI Training Strategies}
The training of generative AI models typically follows a two-stage paradigm: pre-training and fine-tuning. The pre-training stage serves as the foundation, enabling models such as large language models (LLMs) to acquire general language understanding and generation capabilities through exposure to large-scale corpora. The effectiveness of this stage hinges on factors such as the scale and quality of the training data, the design of model architectures, and the implementation of efficient optimization and acceleration techniques. Following pre-training, the fine-tuning stage aims to adapt the model to specific tasks or align it more closely with human preferences. In particular, reinforcement learning from human feedback (RLHF) has emerged as a key strategy for improving the alignment between model outputs and human expectations in terms of helpfulness, factual accuracy, and safety. This approach has been successfully employed in systems such as InstructGPT, Sparrow, and ChatGPT~\cite{ouyang2022training, glaese2022improving}, where models are refined using reward signals derived from human evaluations. Despite the capabilities acquired through large-scale training, these models may still produce outputs that are misaligned with user intent, highlighting the importance of post-hoc alignment techniques like RLHF.

\subsection{Generative AI Pre-training Strategies}
Pre-training serves as the critical foundation for establishing the core capabilities of generative AI models. In this phase, large language models (LLMs) are exposed to massive text corpora through unsupervised or self-supervised learning, enabling them to acquire a broad understanding of linguistic patterns, semantics, and factual world knowledge without the need for explicit task supervision. Common pre-training objectives include autoregressive language modeling--where the model predicts the next token given the previous context (e.g., GPT series)--and masked language modeling, which involves predicting randomly masked tokens within an input sequence (e.g., BERT~\cite{kenton2019bert}). Some model architectures adopt hybrid objectives to enhance versatility and robustness across tasks. The performance of pre-trained models is closely tied to the scale, diversity, and quality of the training data, as well as to the design of the underlying model architecture. 

Transformer-based architectures have emerged as the most prevalent choice for pre-training due to their inherent scalability and parallelizability. To ensure efficient and stable training of billion- or even trillion-parameter models, various optimization algorithms such as Adam and its variants are employed, along with advanced training strategies including mixed-precision training, gradient checkpointing, and curriculum learning. Additionally, system-level techniques--such as pipeline parallelism~\cite{huang2019gpipe}, tensor model parallelism~\cite{shoeybi2019megatron}, and distributed training frameworks--further accelerate the process and make large-scale model training feasible. The result of this stage is a general-purpose model capable of supporting diverse downstream tasks via fine-tuning or prompting, underscoring the essential role of pre-training in the generative AI pipeline.

\subsection{Prompt-learning for Generative AI}
Prompt-learning has become a central mechanism for task adaptation in generative AI, especially for LLMs. It encompasses a range of techniques that allow models to perform new tasks by conditioning on carefully designed inputs, often without additional parameter updates.

The most basic form, manual or automated prompt design~\cite{liu2022design}, aims to craft effective instructions or task templates to elicit desired behaviors from LLMs. These prompts can be generated through heuristic rules or optimized using gradient-based or black-box methods. A more advanced form is In-Context Learning (ICL)~\cite{brown2020language}, introduced with GPT-3, where the model is prompted with a task description and a few exemplars formatted in natural language. ICL enables zero-shot or few-shot generalization by allowing the model to infer task structure from the context without gradient updates. Building upon ICL, Chain-of-Thought (CoT) prompting~\cite{wei2022chain} extends the prompt format by explicitly incorporating intermediate reasoning steps between the input and the expected output. This strategy significantly enhances performance on complex reasoning tasks such as arithmetic, commonsense, and symbolic reasoning. By decomposing problems into sequential steps, CoT provides interpretability and improves task success rates. Recent research continues to explore when CoT prompting is most effective and how to construct such prompts systematically.

\section{AI-generated Content and Detection 
}\label{sec:detection} 

\begin{figure}
    \centering
    \includegraphics[width=0.8\linewidth]{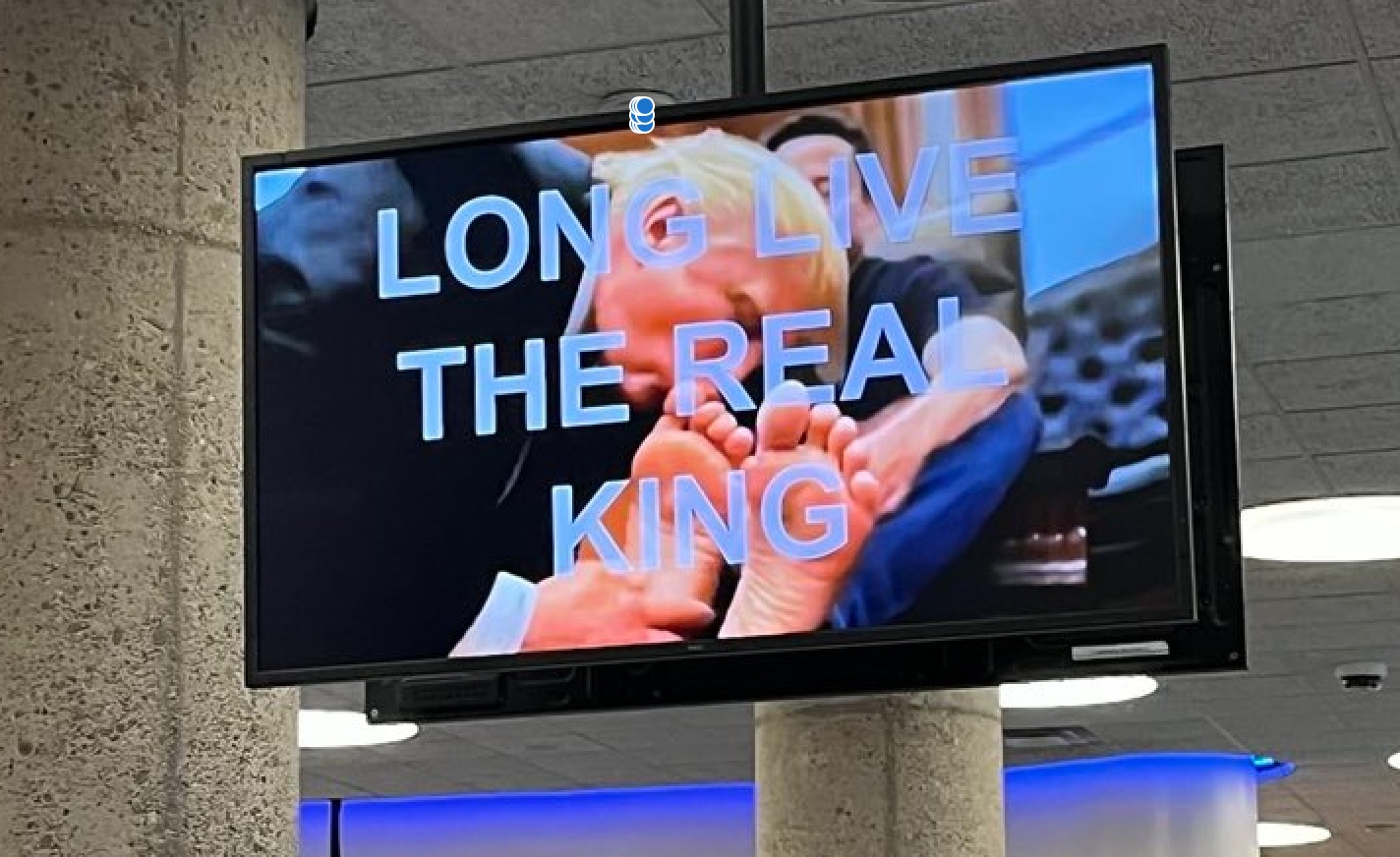}
    \caption{AI-Generated Video Of Trump Kissing Elon Musk’s Feet Reportedly Plays in HUD Building (Posted on The New York Times~\cite{Trump})}
    \label{fig_obj}
\end{figure}

\noindent The rise of \textbf{large language models (LLMs)}, particularly those based on the transformer architecture, represents a fundamental paradigm shift in the creation and dissemination of digital information~\cite{loth2024blessing}. Today, AI can cheaply generate human-like \textbf{text, images, and videos} at scale, making disinformation easier to produce but harder to detect (e.g. Figure~\ref{fig_obj}). False information spreads up to six times faster than truthful news~\cite{su2023adapting}, yet current detection methods struggle to keep pace. This technological leap has transformed the digital landscape, presenting unprecedented challenges to information integrity, public trust, and social stability. The ease with which malicious actors can now generate convincing but false narratives at scale necessitates a deeper, more structured understanding of this new content ecosystem and the threats it contains.

Traditional fake news detectors largely focus on a single modality (typically text) and assume inputs are human-written~\cite{su2023adapting}. Today, AI can cheaply generate human-like text, images, and videos at a scale previously unimaginable, making sophisticated disinformation easier to produce but substantially harder to detect~\cite{loth2024blessing, chen2023llmmisinfo}.
This leaves critical \textbf{gaps and research questions:} how to recognize AI-generated fake articles that mimic legitimate journalism, how to catch doctored images or deepfake videos paired with misleading text, and how to do so in real-time before disinformation proliferates.
Moreover, adversaries continually adapt content to evade detection, creating an arms race between attackers and defenders~\cite{loth2024blessing}. The \textbf{intersection of generative AI and fake news} remains under-explored, and there is a pressing need for research that bridges this gap with comprehensive, multi-dimensional solutions.

\subsection{Overview of Existing Studies}
Prior efforts have addressed portions of the fake news challenge, but no existing work fully targets the new wave of AI-generated disinformation across all modalities. Early research predominantly focused on detecting human-written, text-based fake news, developing models to identify linguistic patterns~\cite{hamed2023review,balshetwar2023fake} and propagation dynamics on social media~\cite{zhou2023linguistic,mahyoob2020linguistic}. A key goal was the early detection of fake news before it could spread widely, often by leveraging the credibility of users and publishers as weak supervisory signals~\cite{wei2021towards,meyers2020fake,yuan2020early}. These foundational works confirmed the severity of the fake news problem and offered innovative detection approaches within their specific scopes. However, the paradigam has shifted. The advent of powerful, accessible generative models necessitates a move beyond these initial efforts, as they were not designed to handle the unique statistical artifacts of AI-generated content or the complexities of multimodal media. Consequently, none have yet delivered a unified solution spanning text, image, and video modalities in the post-LLM era.

Numerous recent research works have tackled pieces of the fake news puzzle in isolation~\cite{wu2024unified,DBLP:conf/kdd/ZhuGYLK24,ma2024event,DBLP:conf/acl/ZongZLL0024,DBLP:conf/acl/LiuWLL24,DBLP:journals/tkde/ZhangZZZWYL25,hu2024bad}. 
This has led to a fragmented landscape of specialized solutions, including frameworks for unified evidence enhancement~\cite{wu2024unified}, propagation-aware graph transformers~\cite{zhou2024edu}, and event-driven multimodal learning~\cite{ma2024event}. Other specialized approaches focus on detecting opinion evolution in short videos~\cite{DBLP:conf/acl/ZongZLL0024}, generating reinforced propagation paths for early detection~\cite{DBLP:journals/tkde/ZhangZZZWYL25}, and using LLMs as advisors to smaller models~\cite{hu2024bad}.

To our knowledge, no existing system provides a single fake news detector that integrates all three modalities and is explicitly designed to handle AI-generated content and adaptive adversarial attacks. 

\subsection{Research Challenges}
As generative and detection technologies co-evolve, the research community is grappling with increasingly sophisticated challenges that push the boundaries of current defense mechanisms. These frontier challenges are moving beyond simple binary classification (human vs. AI) toward more nuanced, context-aware, and veracity-focused problems.

\noindent
\textbf{Adversarial Robustness in Co-evolutionary Arms Race.}
The primarily challenge is ensuring adversarial robustness in the face of an escalating, endogenous arms race. A particularly potent threat is the rise of style-conversion attaks, where a malicious actor uses an LLM to rewritten fake news in the distinct linguistic style of a trusted source like \textit{The New York Times}. This technique is designed to fool detectors that rely on identifying the stylistic artifaxts of AI-generated text. The SheepDog framework~\cite{wu2024sheepdog} was developed specifically to counter this threat by training a detector to be style-agnostic, thereby forcing it to focus on substantive content rather than superficial style. This cat-and-mouse dynamic requires the development of detection systems that are not only accurate but also resilient to adaptive adversaries.

\noindent
\textbf{The Multimodal Imperative: Beyond Textual Analysis.}
While text-based disinformation remains a threat, the next major battleground is unequivocally multimodal, involving hyper-realistic deepfake videos, manipulated images paired with misleading captions, and deceptive short-form video content. Detecting multimodal disinformation is inherently more complex than text-only analysis, requiring a multi-level verification process. Disinformation campaigns now routinely involve hyper-realistic deepfake videos, manipulated images paired with misleading captions, and deceptive short-form video content. The academic community has responded by creating numerous large-scale datasets for detecting these manipulations, such as FakeAVCeleb, Celeb-DF, and DeeperForensics-1.0~\cite{akhtar2023mmfactcheck,guo2022afcsurvey,yao2023mmfc}. Detecting multimodal disinformation is inherently more complex than text-only analysis. It requires a multi-level verification process that not only assesses the authenticity of each individual modality but also, verifies the semantic consistency between them (e.g., does the image actually depict the events described in the caption?).

\noindent
\textbf{The Hallucination Problem: Distinguishing Lies from Errors.}
A distinct and subtle challenge is distinguishing deliberate, malicious disinformation from unintentional factual errors, or ``hallucinations,'' which are an inherent limitation of current LLM architectures. While a style-conversion attack is an act of deception, a hallucination is a system failure. Detecting these unintentional falsehoods requires different techniques. Black-box approaches like SelfCheckGPT operate on the premise that if an LLM is confident about a fact, its statements will be consistent across multiple, stochastically sampled responses~\cite{manakul2023selfcheckgpt}; if it is hallucinating, the responses will likely diverge and contradict one another~\cite{muhammed2025selfcheckagent}. SAPLMA have shown that a model's internal activation patterns can be used to train a classifier that predicts, which high accuracy, whether the model is generating a factual statement or hallucinating~\cite{park2025steer, azaria2023internal}. Addressing this problem is critical for building trustworthy AI systems, as both intentional and unintentional falsehoods erode public trust.

\subsection{Research Propositions}
To address the frontier challenges, future research should prioritize the development of more robust, context-aware, and holistic detection frameworks. Several promising research propositions have emerged from recent works.

\noindent
\textbf{LLMs as Collaborative Analyzers: The ``Bad Actor, Good Advisor'' Paradigm.}
Empirical studies have shown that while LLMs, when used in a zero-shot capacity, often under perform smaller, fine-tuned models at the final task of fake news classification, they excel at generating nuanced, multi-perspective rationales and leveraging commonsense knowledge~\cite{hu2024bad}. This observation led to the ``Bad Actor, Good Advisor'' paradigm, where the LLM is not the final decision-maker but an advisor to a more specialized model. The \textbf{ARG} framework exemplifies this approach by using an LLM to generate explanatory rationales for a new article from various perspectives (e.g., style, factuality, logic)~\cite{hu2024bad}. To overcome this limitations, one way is to use LLMs to transform into autonomous agents \cite{hosseini2025role} that actively seek and synthesize external information. This approach mimics the process of human fact-checkers who consult various sources to verify a claim. 

\noindent
\textbf{LLMs as Autonomous Tool-Using Agents.}
To overcome the limitations of their static, internal knowledge \cite{liu2025knowledge,li2025hycube}, LLMs can be transformed into autonomous agents that actively seek and synthesize external information. This approach mimics the process of human fact-checkers who consult various sources to verify a claim. The FacTool Framework provides a general architecture for this, where an LLM orchestrates a multi-step process: it first extracts verifiable claims from a text, then generates appropriate queries for external tools, and finally reasons over the collected evidence to verify the claim~\cite{chern2023factool}. LEMMA framework is a powerful multimodal instantiation of this paradigm, specifically designed for verifying image-text pairs by autonomously using tools like reverse image search to trace an image's origin and context~\cite{xuan2024lemma}. This directly addresses the multimodal challenge by grounding the model's reasoning in external, verifiable evidence.

\noindent
\textbf{LLMs as Interpretable Reasoners.}
A critical weakness of many deep learning-based detector is their ``black box'' nature, which undermines trust and makes it difficult to diagnose failures. To improve reasoning and interpretability, a new wave of research aims to make the LLM's process transparent. Instead of a simple debate, frameworks are being developed to structure this reasoning.  For instance, ProgramFC guides an LLM to decompose a complex fact-checking task into a sequence of simpler, logical steps formatted as a computer program, making the entire verification process auditable~\cite{pan2023fact}. Similarly, TELLER employs a dual-system architecture inspired by human cognition. Its ``cognition system'' uses an LLM to answer a set of simple, human-defined logical questions about a news article~\cite{liu2024teller}. A separate, symbolic ``decision system'' then aggregates these atomic logical answers using learnable rules to derive a final, explainable verdict. This hybrid approach ensures that the final decision is based on an explicit and generalizable logical foundation, moving beyond simple classification to trustworthy, reasoned verification.

\subsection{Concluding statement}
The dynamic interplay between AI-driven content generation and detection has created an escalating, endogenous arms race. As generative models grow more sophisticated, the focus of detection is necessarily shifting from superficial stylistic analysis towards deeper, content-based, and reasoning-driven paradigms. Looking forward, the most pressing challenges lie at the intersection of adversarial robustness and multimodality, demanding solutions that are resilient to stylistic manipulation and capable of reasoning across different data types. Future research should prioritize the development of hybrid, interpretable, and multimodal detection systems. A promising path forward involves creating integrated frameworks that combine the cognitive power of LLMs with the efficiency of specialized models trained for style-agnostic content analysis and rigorous cross-modal consistency checking. Only through such a multi-pronged, resilient approach can the research community hope to maintain information integrity in an increasingly synthetic media landsape.

\section{Spread and Use of AI-generated Information 
} \label{sec:spread} 

\subsection{Overview of Existing Studies}

The spread of AIGC is driven by the same digital distribution infrastructures that drive human-created content: social media feeds, search engines, recommender systems \cite{yuan2024hide,chen2024explainable,li2024session}, and syndication networks, but operates at a greater scale and speed due to its low marginal production cost and ability to be instantly re-prompted and re-purposed \cite{loth2024blessing,chen2023llmmisinfo}.
Empirical studies of online information diffusion show that algorithmic curation systems amplify engaging content regardless of provenance, enabling AI-generated text, images, and videos to achieve viral reach within hours of publication, often crossing platforms via automated reposting and embedding \cite{loth2024blessing,guo2022afcsurvey}.
Controlled experiments confirm that synthetic news articles and deepfake media can propagate more rapidly than equivalent human-produced material because of their optimized linguistic fluency and visual realism, which increase click-through and sharing rates \cite{loth2024blessing,chen2023llmmisinfo}.
This accelerated diffusion extends beyond misinformation: in marketing, corporations employ AIGC in programmatic advertising and personalization engines to micro-target audiences, with generated copy and imagery dynamically tailored to demographic or psychographic profiles in real time \cite{cillo2024generative,reis2022contentmarketing,chen2025ctr}.
In journalism, early adoption has leveraged newswire-style syndication and automated social media posting to disseminate AI-generated summaries, headlines, and visuals across multiple outlets and platforms almost instantaneously \cite{yao2023mmfc,akhtar2023mmfactcheck,liu2024teller}.
These sector-specific pipelines ensure that once an AI-generated asset is created, it can be seamlessly integrated into high-volume content streams, such as marketing email blasts, push notifications, and auto-populated website feeds, maximizing both reach and repeated audience exposure \cite{cillo2024generative,reis2022contentmarketing}.

\subsection{Research Challenges}\label{sec:spread_research_challenges}

As generative and dissemination technologies rapidly advance, the research community is confronting increasingly complex challenges in monitoring and managing the spread of AI-generated information. These frontier issues extend beyond detecting whether content is AI- or human-generated, shifting toward understanding its cross-platform amplification dynamics, preserving provenance through transformation, mitigating cascade effects of inaccuracies, and ensuring equitable safeguards across languages and sectors. Addressing these challenges requires an integrated view of both technical vulnerabilities and the socio-technical systems through which AI-generated information circulates.

\textbf{Rapid, algorithmically mediated amplification.} Across platforms, ranking and recommendation functions surface high-engagement items regardless of provenance, allowing AI-generated text, image, audio, and video to achieve fast, large-scale reach once they trigger feedback loops of clicks, shares, and watch time~\cite{loth2024blessing,sharma2022misinfo}. Comparative audits and user studies indicate that AI-written fabrications can be more persuasive than human-written falsehoods due to higher linguistic fluency and coherence, widening the window in which synthetic items accumulate engagement before moderation or fact-checking intervenes~\cite{chen2023llmmisinfo,loth2024blessing}. Cross-platform diffusion further compounds speed and scale: content routinely migrates from source communities to mainstream feeds and news blogs, fragmenting provenance and frustrating coordinated takedowns across ecosystems~\cite{horta2021crossplatform,sharma2022misinfo}.

\textbf{Detection brittleness and adversarial adaptation.} Text-origin detectors show substantial accuracy drops under simple paraphrasing, translation, or style transfer, undermining reliability in the wild~\cite{uchendu2021turingbench,krishna2023paraphrase,liang2023detectorbias}. Methodological advances such as DetectGPT improve zero-shot detection by exploiting curvature of the log-likelihood landscape, yet performance still degrades under distribution shift and post-editing~\cite{mitchell2023detectgpt,uchendu2021turingbench}. Adversaries can actively optimise for evasion, e.g., re-writing in the style of trusted outlets (style-conversion attacks) or passing content through multiple model pipelines to erase detectable artefacts~\cite{wu2024sheepdog,krishna2023paraphrase}.

\textbf{Multimodal forensics under transformation.} Deepfake detectors trained on canonical manipulations lose robustness under benign transformations (compression, resizing) and adversarial perturbations that preserve human-perceived semantics~\cite{neekhara2021adversarialdeepfakes,sharma2022misinfo}. While high-quality datasets (e.g., Celeb-DF v2; DeeperForensics-1.0) catalysed progress, generalisation across creators, generation engines, and editing toolchains remains limited, leading to domain shift failures in the wild~\cite{li2020celebdf,jiang2020deeperforensics}. For image models, recent watermarking in diffusion latent spaces (e.g., tree-ring signatures) offers promising provenance signals, but robustness to cropping, re-generation, or screenshot-to-text pipelines is still an open challenge~\cite{wen2023treering,sharma2022misinfo}.

\textbf{Provenance, watermarking, and survivability.}
Cryptographic and statistical watermarking methods for LLM outputs offer the potential to signal content provenance at generation time. However, existing approaches face fundamental trade-offs between detectability, utility, and robustness to transformations such as paraphrasing or summarization~\cite{kirchenbauer2023watermark,carlini2024watermarklimits}. In heterogeneous content ecosystems, metadata-based provenance markers are frequently lost during reposting, format conversion, or platform-side re-encoding, breaking audit trails precisely when cross-platform dissemination is most rapid~\cite{sharma2022misinfo,kirchenbauer2023watermark}.

\textbf{Hallucination, calibration, and cascade effects.}
Even without malicious intent, models may generate fluent but factually incorrect statements. While black-box self-consistency checks and white-box activation-based signals can improve calibration, they remain insufficiently reliable for deployment at platform scale~\cite{manakul2023selfcheckgpt,ji2023hallucinationsurvey}. Once published, such errors can be quoted, remixed, and algorithmically amplified, both by recommender systems and downstream AIGC pipelines, transforming local inaccuracies into widely propagated pseudo-facts~\cite{sharma2022misinfo,ji2023hallucinationsurvey}.

\textbf{Bias, equity, and linguistic coverage.} Detectors can be biased against non-native writing styles, increasing false-positive risks for human authors and complicating equitable moderation policies~\cite{liang2023detectorbias}. Multilingual and low-resource settings remain under-served in both generation safeguards and detection tooling, enabling asymmetric spread where monitoring capacity is weakest~\cite{sharma2022misinfo,uchendu2021turingbench,zhao-aletras-2024-comparing}.

\textbf{Governance and operationalisation gaps.} Sectoral policies (e.g., editorial standards, claims-substantiation) have not yet translated into measurable reductions in undesirable spread, in part because organisations lack validated, cross-platform APIs and auditing protocols for AIGC dissemination~\cite{loth2024blessing,sharma2022misinfo}. End-to-end evaluations that couple provenance, detection, and intervention timing with downstream audience impact are scarce, leaving open what combinations of technical and procedural controls actually curb harmful propagation without suppressing beneficial uses~\cite{akhtar2023mmfactcheck,loth2024blessing}.

\subsection{Research Propositions}

To address the frontier challenges identified in Section~\ref{sec:spread_research_challenges}, future research should prioritise the development of spread-aware, provenance-preserving, and cross-platform-resilient frameworks for AI-generated information governance. Several promising research propositions can be drawn from recent works.

\textbf{Quantifying spread velocity differentials.} Empirical studies should systematically compare the dissemination velocity of AI-generated content against human-generated counterparts across different platforms and domains, isolating the role of algorithmic amplification, content fluency, and visual realism in accelerating reach~\cite{loth2024blessing,chen2023llmmisinfo,sharma2022misinfo}. Such analyses can reveal the precise speed advantage AIGI gains, enabling better calibration of detection and intervention timing.

\textbf{Platform-specific spread modeling.} The spread patterns of AIGI are likely to differ between infrastructures such as news syndication networks, social media recommender systems, and targeted advertising ecosystems~\cite{cillo2024generative,yao2023mmfc,horta2021crossplatform}. Building predictive models that capture these structural differences will inform tailored intervention strategies that balance content utility with harm mitigation.

\textbf{Provenance-preserving generation and transmission.} Embedding transformation-resilient provenance data at the point of generation, and maintaining it through cross-platform sharing, could reduce unattributed or malicious spread~\cite{kirchenbauer2023watermark,takale2024watermark,lancaster2023watermark}. Research should evaluate robustness against common transformations (e.g., paraphrasing, cropping, re-encoding) and design watermarking or cryptographic techniques that persist without degrading user experience.

\textbf{Multimodal, context-aware detection in the spread loop.} Detection tools should be integrated into dissemination pipelines (e.g., pre-publication checks, syndication feeds) and account for multimodal cues, cross-platform reformatting, and context-specific risk profiles~\cite{akhtar2023mmfactcheck,wen2023treering,neekhara2021adversarialdeepfakes}. This shifts detection from a post-hoc, single-platform task to a proactive, distributed safeguard.

\textbf{Sectoral spread signature mapping.} Mapping the ``spread signatures'' of AIGI in different sectors, e.g., journalism, marketing, grassroots public use, can reveal distinctive channel preferences, audience re-sharing behaviours, and amplification dynamics~\cite{cillo2024generative,liu2024teller,reis2022contentmarketing}. This knowledge can underpin sector-specific policy guidelines and evaluation metrics for both beneficial and harmful dissemination.

\subsection{Concluding statement}
Recent academic work indicates that the spread of AIGI is inseparable from the infrastructures that mediate modern information flows, in many cases exploiting the same programmatic, syndication, and algorithmic systems that make digital communication efficient \cite{loth2024blessing,cillo2024generative}.
While this enables beneficial uses, such as personalized marketing, timely news summaries, and creative cultural production, it also amplifies the risks of rapid misinformation proliferation, provenance loss, and the amplification of low-quality or manipulative content.
Addressing these issues requires both technical interventions (provenance-preserving formats, cross-platform detection) and policy or governance measures tailored to the distinct spread dynamics in different sectors \cite{lancaster2023watermark,liu2024teller}.

\section{Impact of AI-generated content to public trust 
}\label{sec:publictrust} 
Trust in information is not merely a psychological sentiment; it is a structural pillar supporting democratic governance, economic stability, social cohesion, and public health. In the digital era, the channels through which information flows, such as, social media platforms, search engines, content aggregators, have already transformed how trust is built and maintained. With the advent of AI-generated content (AIGC), this trust ecosystem is undergoing a fundamental shift. 

AIGC refers to text, images, audio, and video produced wholly or partially by algorithms, often indistinguishable from human-authored material. Modern generative AI systems, such as large language models (LLMs) and diffusion-based image generators, can produce outputs at unprecedented speed, scale, and stylistic fidelity~\cite{loth2024blessing}. This technical leap brings new opportunities for personalized communication, scalable knowledge dissemination, and creative augmentation. But, it also erodes the traditional cues audiences use to evaluate the credibility of content.

From a sociotechnical perspective, public trust is shaped by three interacting dimensions:
\begin{itemize}
    \item Content Authenticity: perceived truthfulness, originality, and absence of manipulation.
    \item Source Credibility: perceived reliability of the entity disseminating the content.
    \item Transparency and Accountability: clarity around how, why, and by whom content was generated.
\end{itemize}

When AI systems become primary content creators, these dimensions are destabilized. Provenance is harder to verify, attributions may be obscured, and “authorship” becomes a diffuse concept encompassing algorithms, data curators, and human prompt designers.

The trust challenges posed by AIGC are amplified by its integration into high-velocity dissemination channels. Rumor theory and empirical misinformation research show that speed of propagation is inversely correlated with verification likelihood~\cite{chen2020rumor}. AI not only accelerates rumor spread but also enables content mutation, producing multiple stylistic variants that evade detection algorithms~\cite{akhtar2023mmfactcheck}. In this sense, AIGC acts as both a catalyst and a chameleon in the rumor ecosystem.

Crucially, the trust impact of AIGC is domain-specific:
\begin{itemize}
    \item In journalism, the risk is epistemic trust erosion-audiences doubt the reliability of entire media outlets once exposed to AI-generated misinformation, even if most reporting is accurate~\cite{zhou2025effect}.
    \item In politics, micro-targeted AI propaganda can trigger selective trust-audiences believe narratives aligned with prior attitudes while dismissing others as artificial manipulation~\cite{battista2024political}.
    \item In public health, trust operates as a life-critical filter-AI-generated inaccuracies can reduce adherence to medical guidance, undermining institutional credibility~\cite{quinn2021trust}.
    \item In education, over-reliance on AI tutors may cultivate functional trust (confidence in task completion) but diminish relational trust (confidence in ethical and human-centered judgment)~\cite{zhai2024effects}.
\end{itemize}
The emerging picture is one of a double-edged trust dynamic:
\begin{itemize}
    \item Trust Augmentation occurs when AI delivers accurate, transparent, and contextually relevant information, especially in resource-constrained environments where human-generated content is unavailable or delayed.
    \item Trust Degradation occurs when AI introduces inaccuracies, obscures provenance, or is perceived as intentionally manipulative-effects that can spill over to unrelated human-authored content, creating a generalized skepticism toward all media~\cite{schiff2020liar}.
\end{itemize}
This complexity emphasises the need for trust-by-design frameworks-systems that embed provenance tracking, bias mitigation, and context-sensitive disclosure into the AIGC production pipeline. Without these safeguards, the very capabilities that make AIGC powerful risk undermining the social contract between information providers and the public.

\subsection{Overview of Existing Studies}
The literature examining the relationship between AI-generated content (AIGC) and public trust has grown significantly over the past three years, reflecting the accelerated adoption of large language models (LLMs) and generative multimodal systems in high-impact domains. Existing studies reveal a nuanced and often contradictory landscape: while AIGC can improve efficiency, accessibility, and personalization, it can also undermine credibility and amplify skepticism when provenance, accuracy, or intent are unclear.

A recurring finding across domains is that audience trust is highly context-dependent. Trust perceptions vary based on domain sensitivity, audience familiarity with AI, nature of the task, and disclosure practices. For example, AIGC in low-stakes entertainment contexts is often met with curiosity and acceptance, whereas its use in high-stakes settings such as news reporting, political communication, or healthcare tends to trigger scrutiny and potential distrust.
\subsubsection{AI-Generated Content in News Media}
The integration of AIGC into journalism has primarily been studied in two contexts:
\begin{itemize}
    \item [(1)] Automated reporting and summarization: LLMs and NLG systems generate news briefs, financial reports, and event summaries. Multiple controlled experiments~\cite{gilardi2024willingness,gilardi2024disclosure} show that readers often rate AI-authored news as comparable to human-authored news in terms of clarity, grammatical quality, and informational density.
    \item [(2)] Synthetic visual and audiovisual content: Deepfake videos, AI-generated imagery, and voice clones are increasingly being deployed in political and social contexts, often without adequate disclosure~\cite{romero2025deepfake,romero2024generative}. Exposure to such manipulated media can cause epistemic trust erosion, where audiences begin to doubt even genuine journalistic content.
\end{itemize}

A key challenge identified is perceptual indistinguishability: when AI-generated news is sufficiently human-like, audiences cannot reliably differentiate it from human work without explicit labeling~\cite{hadan2024great}. However, labeling itself presents a disclosure paradox-disclosed AI authorship can lower perceived credibility~\cite{wittenberg2025labeling} even for factually accurate pieces, while undisclosed authorship risks ethical and reputational harm if later revealed.

\subsubsection{Political and Social Impacts}
Political communication has emerged as one of the most sensitive arenas for AIGC trust research. Studies by Islam et. al.~\cite{islam2024ai} and Romanishyn et. al.~\cite{romanishyn2025ai} document how generative AI can be exploited to create hyper-targeted propaganda-highly tailored narratives designed to resonate with specific voter demographics. These narratives leverage psychographic profiling and microtargeting algorithms to maximize persuasion efficiency.

One particularly concerning dynamic is belief persistence: once individuals adopt a belief from AI-generated political misinformation, subsequent corrections-whether human- or AI-provided-often fail to fully restore prior trust levels~\cite{duan2025understanding,sanna2025belief}.
Moreover, the liar’s dividend effect is increasingly visible in political discourse: awareness that content can be faked by AI provides plausible deniability to political actors, enabling them to dismiss authentic evidence as fabricated.

\subsubsection{Health Information and AI-Generated Content}
The healthcare domain presents a double-edged trust trajectory. On one side, AI-driven health chatbots, virtual assistants, and medical writing tools can enhance access to credible health information, especially in underserved regions~\cite{bajwa2021artificial}. On the other, studies reveal that even minor factual errors in AI-generated health content can severely damage trust-not only in the AI system but in the broader healthcare institution endorsing it~\cite{quinn2021trust}.

Recent literature identifies three interdependent factors that shape whether patients and the public develop or erode trust in AI-driven health communication: transparency, explainability, and consistency.
\begin{itemize}
    \item [1.] Transparency: Users must be informed when health information is generated by AI. Disclosure reduces perceptions of deception and aligns with ethical standards, though some studies show it can lower perceived credibility even when content is accurate~\cite{longoni2019resistance,chen2021impacts,naumova2023responsible}. This creates a disclosure paradox: necessary for ethics, but potentially harmful to trust if not paired with assurances of professional oversight.
    \item [2.] Explainability: Trust increases when systems provide reasoning steps or cite reputable sources, rather than offering opaque answers. Experiments with health chatbots show that explainable responses with references to medical guidelines score higher in credibility among both patients and clinicians~\cite{amann2020explainability,ayers2023comparing,nadarzynski2019acceptability}.
    \item [3.] Consistency: AI outputs must align with professional standards. Even small errors or “hallucinations” can damage trust and lead to spillover skepticism toward otherwise correct content~\cite{haverkamp2023chatgpt}. When AI advice is explicitly bench-marked against clinical protocols, user confidence improves significantly~\cite{kumthekar2025second,nadarzynski2019acceptability}.
\end{itemize}

In mental health contexts, preliminary research shows that some patients trust AI's non-judgmental tone, but most prefer hybrid systems where AI augments rather than replaces human professionals~\cite{jarrahi2022artificial}.

\subsubsection{Education and Academic Integrity}
Within education, AIGC has been applied to intelligent tutoring systems, automated grading, and personalized learning content~\cite{wang2025research,cheng2024impact}. Early adoption studies show that students generally trust AI tutors for factual and procedural tasks (e.g., solving mathematical problems, debugging code, or generating practice questions), where outputs can be objectively verified. However, skepticism arises when AI is expected to provide nuanced feedback in areas such as essay writing, critical thinking, or moral reasoning, where human judgment and contextual sensitivity are essential~\cite{kizilcec2024perceived}.

Concerns about academic integrity are particularly acute. The widespread availability of generative models such as ChatGPT has intensified fears of plagiarism, ghostwriting, and over-reliance on AI-generated assignments, raising questions about skill mastery and genuine learning outcomes~\cite{lodge2024evolving,bin2023use}. Several studies have noted a paradox: while students may use AI tools for efficiency, they simultaneously express doubts about the fairness of AI detection systems, fearing false accusations of misconduct~\cite{huang2025evaluating}.

Equally significant are issues of assessment transparency and fairness. Institutions face reputational risks if AI-assisted grading is perceived as opaque, inconsistent, or biased~\cite{dwivedi2023opinion,cotton2024chatting}. Trust in educational institutions therefore hinges on ensuring that AIGC is deployed with:

\textbf{Clear evaluation criteria}: AI-supported grading must be explicitly aligned with well-defined rubrics and learning outcomes. This ensures that students perceive assessments as consistent, predictable, and fair, rather than arbitrary outputs of an opaque system. Establishing benchmarked rubrics also provides a reference point for appeal processes when disagreements arise.

\textbf{Human oversight}: While AI can accelerate grading and feedback cycles, ultimate responsibility must remain with educators. Incorporating teachers in the review loop allows for correction of misclassifications, contextual interpretation of student work, and assurance that assessments account for creativity and originality-dimensions AI often struggles to evaluate reliably.

\textbf{Transparent communication}: Trust is strengthened when students are informed about the extent and purpose of AI use in evaluation. Transparency involves disclosing not only \textit{when} AI is used (e.g., in preliminary grading or plagiarism detection), but also \textit{how} decisions are made and \textit{why} they align with institutional policies. Clear communication helps mitigate perceptions of bias or unfairness, reinforcing institutional accountability.

Emerging research suggests that trust can be strengthened by framing AI as a collaborative assistant rather than a replacement for educators, where human teachers retain authority over final judgments~\cite{lee2024impact,cheng2024impact}. This hybrid approach not only addresses integrity concerns but also reassures students that educational institutions remain committed to fairness, accountability, and the human dimensions of learning.

\subsection{Research Challenges}

Despite rapid adoption of AI-generated content (AIGC) across domains, several challenges remain unresolved in maintaining and strengthening public trust:

\textbf{Transparency and Disclosure}: Current disclosure practices are inconsistent. While some organisations openly label AI-assisted content, others obscure its origin. Determining the appropriate level and form of disclosure without undermining consumer confidence is a key research challenge.

\textbf{Bias and Fairness}: Generative models inherit biases from training data, raising risks of discriminatory or culturally insensitive outputs. Identifying, quantifying, and mitigating these biases remains an open research problem.

\textbf{Authenticity versus Efficiency Trade-off}: AIGC enables scalable production of personalised content, but excessive automation can erode perceptions of authenticity. Research must explore how to balance efficiency with the need for human-like connection.

\textbf{Evaluation Metrics for Trust}: Traditional performance metrics (e.g., accuracy, BLEU scores) fail to capture socio-psychological aspects of trust. Developing robust, multi-dimensional trust evaluation frameworks for AIGC is essential.

\textbf{Regulatory and Ethical Alignment}: Policies such as the EU AI Act and FTC guidelines are emerging, yet there is limited clarity on how organisations should operationalise them. Research must address how compliance can be embedded into AI pipelines without stifling innovation.

\subsection{Research Propositions}

Building on the identified challenges, this paper proposes several avenues for future research:

\textbf{Trust-Centric Design of AIGC Systems}: Move beyond accuracy-driven models towards systems optimised for perceived transparency, fairness, and explainability. This includes embedding disclosure cues, rationale explanations, and fairness constraints directly into generation processes.
    
\textbf{Hybrid Human--AI Communication Models}: Investigate frameworks where human oversight complements AI automation, particularly in high-stakes or emotionally sensitive contexts. This approach can leverage AI’s scalability while preserving human empathy and accountability.
    
\textbf{Trust Measurement Frameworks}: Develop interdisciplinary methodologies to evaluate trust in AIGC, combining behavioural experiments, longitudinal surveys, and computational trust indices. Such frameworks should capture both cognitive (e.g., reliability) and affective (e.g., authenticity) dimensions.
    
\textbf{Cross-Cultural Studies of Trust in AIGC}: Trust perceptions vary across cultural and demographic groups. Comparative research can uncover context-specific factors that shape acceptance or skepticism, enabling more inclusive AI practices.
    
\textbf{Governance and Standardisation Pathways}: Explore how industry standards, auditing protocols, and certification schemes can provide external guarantees of trustworthiness, similar to established practices in cybersecurity and data privacy.

\subsection{Concluding Statement}

The rise of AI-generated content represents a pivotal shift in how information is created, disseminated, and consumed. While AIGC offers unprecedented opportunities for personalisation, efficiency, and scale, its uncritical deployment risks undermining the very foundation of digital engagement: public trust. This opinion paper has argued that the trajectory of AIGC adoption will be determined not merely by technical performance but by the ability of researchers, practitioners, and policymakers to address transparency, fairness, and authenticity in content generation. By reframing AIGC as a trust-sensitive technology, future research can ensure that its integration strengthens rather than erodes the social contract between humans and digital systems.

\section{Impact of AI-generated Content to Digital Marketing
}\label{sec:digitialmarkteting} 

Digital marketing \cite{chaffey2019digital,bala2018critical} has become a central component of modern business strategies, driven by the rapid growth of online platforms and digital communication channels. Companies increasingly leverage tools such as social media, search engines, and content marketing to reach and engage targeted customers. 

With the rise of large language models (LLMs) and improvements in generative AI, AI-generated content (AIGC) is becoming an increasingly important asset in digital marketing \cite{nalbant2025bibliometric,reed2025artificial,cillo2024generative}. Powered by massive pre-trained LLMs, AIGC systems can produce diverse types of marketing content, including text, images, audio, and videos, that are tailored to specific customers. Such capabilities enable marketers to rapidly generate personalized content through prompt engineering, thereby reducing reliance on manually created marketing content and improving overall efficiency.

\begin{figure*}
    \centering
    \includegraphics[width=0.8\linewidth]{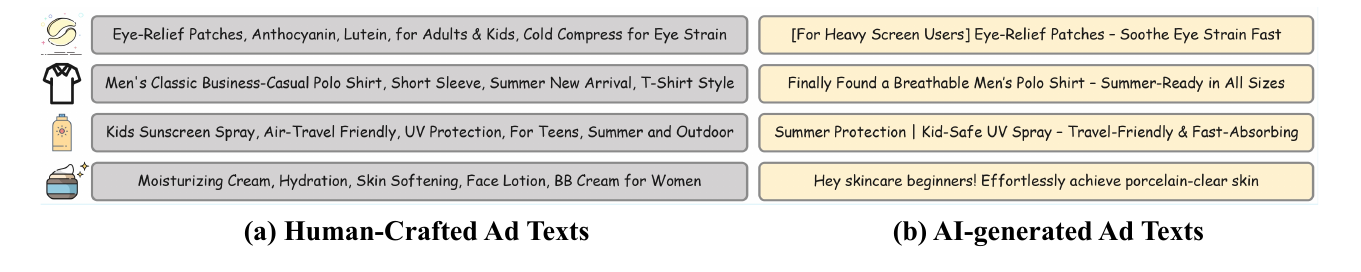}
    \caption{Examples comparing human-crafted ad titles and AI-generated ad titles for online shopping platform \cite{chen2025ctrtext}.}
    \label{fig:itemtitlegeneration}
\end{figure*}
\begin{figure*}
    \centering
    \includegraphics[width=0.8\linewidth]{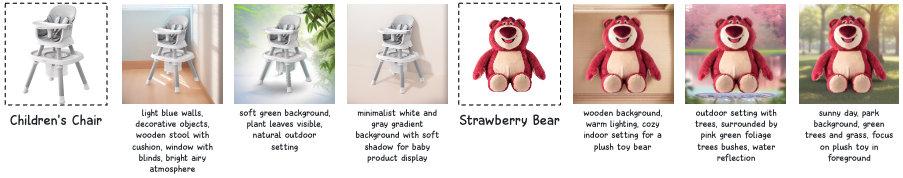}
    \caption{Examples of generating diverse advertising image backgrounds using different prompts \cite{chen2025ctr}.}
    \label{fig:aigeneratedbackground}
\end{figure*}
\subsection{Overview of Existing Studies}
In this section, we categorize existing methods according to the type of generated content, including textual marketing content generation and visual marketing content generation, and introduce representative methods under each category.

\textbf{Textual marketing content generation.} Text content is widely used in digital marketing scenarios such as headlines for advertisements, product titles (as illustrated in Figure \ref{fig:itemtitlegeneration}), and search engine optimization (SEO) content.
For example, In \cite{hughes2019generating}, the generation of SEO text ads is formulated as a sequence-to-sequence task, and the Self-Critical Sequence Training technique is applied to produce ads that are both human-like and more effective.
PLATO-Ad \cite{lei2022plato} utilizes multi-task prompt learning mechanism to generate text ads, and improves click-through rates (CTR) on both search ad descriptions and feed ad titles in real-world online advertising platforms. 
CREATER \cite{wei2022creater} proposes generating ad texts based on high-quality user reviews to improve CTR by contrastive learning technique.
In addition, to address the higher cost and less efficient of traditional human-written SEO content, a semi-automated method leveraging natural language generation (NLG) techniques is introduced in \cite{reisenbichler2022frontiers}, which drafts machine-generated texts refined by human editors. Results indicate that these texts are nearly indistinguishable from expert-written SEO content and achieve superior search engine rankings, significantly improving the cost-effectiveness of content marketing.
Furthermore, to improve CTR on online shopping platforms, a two-step method for generating item titles is proposed in \cite{chen2025ctrtext}. The method first applies in-context learning to produce diverse ad samples, and then refines them through preference optimization based on real online feedback.

\textbf{Visual marketing content generation.} 
Compared to text, another common form of AI-generated content in digital marketing is visual content \cite{hartmann2025power}, such as product image backgrounds (as shown in Figure \ref{fig:aigeneratedbackground}) and promotional banners, which present product features more intuitively and are often more effective in capturing user attention in advertising contexts.
For example, a CTR-driven study \cite{chen2025ctr} employs multimodal large language models to generate advertising image backgrounds based on experts’ prompts. The model is first pre-trained on a large-scale e-commerce multimodal dataset and then fine-tuned through reinforcement learning with a reward model that reflects user click preferences.
In another example, the MARK-GEN framework \cite{islam2024transforming} leverages generative artificial intelligence techniques in the fashion industry. It demonstrates the application of virtual try-on models, such as SVTON \cite{islam2022svton}, to generate clothing product images for use in advertisement banners and on business websites. 
Moreover, the image-to-video model FashionFlow \cite{islam2023fashionflow} is introduced to automatically generate marketing videos from product images. 
Furthermore, a large-scale text-to-image generation framework \cite{zhao2024enhancing} leverages bilingual multimodal alignment to generate semantically coherent advertising images, achieving CTR improvements in real-world applications.

\subsection{Research Challenges}
While existing studies have demonstrated the potential of AI-generated content in digital marketing through various text- and image-based applications, several important research challenges remain.

\textbf{Multi-Modal digital marketing content generation.} Most current AI-generated content for digital marketing methods only focus on single-modality content generation, such as text and images. However, real-world marketing content is often inherently multi-modal, and other modalities still require manual creation by experts. For example, marketing videos need to integrate consistent images, text, and audio, yet how to automatically generate such unified multi-modal content remains largely unexplored. 
Achieving this goal is challenging due to the need for semantic and temporal alignment across modalities, and the requirement for high-quality and controllable outputs suitable for professional marketing.
Although recent advances in multimodal large language models and cross-modal alignment techniques offer promising solutions \cite{yu2025aligning}, affordable and scalable end-to-end frameworks for unified multi-modal marketing content generation remain in their early stages.

\textbf{Personalized content generation.} While AI-generated digital marketing content has the potential to be personalized, current approaches have yet to fully realize this capability. Many existing systems still adopt a “one-size-fits-all” strategy, generating the same content for all users and overlooking specific preferences. For instance, for identical product descriptions or images, some users may prefer a minimalist and elegant style, whereas others may favor different aesthetics. Yet, current methods rarely incorporate fine-grained features such as users’ historical interactions \cite{wang2025generative} and cultural backgrounds, which limits the alignment between generated content and users’ preferred styles \cite{xu2025personalized}, ultimately leading to suboptimal performance.

\subsection{Research Propositions}
Although recent studies have investigated a range of methods for AI-generated content in digital marketing, several key areas remain insufficiently explored.

\textbf{Open benchmarks and datasets.} Although AI-generated content is becoming more common in digital marketing, there are still very few open benchmarks to provide a standardized and fair way to evaluate different methods across various application scenarios. This makes it difficult for researchers and practitioners to compare models, understand their strengths and weaknesses, and track progress in the field. One example is the CAMERA benchmark \cite{mita2023striking}, which focuses on the task of automatic ad text generation (ATG). Another example is the ADPARAPHRASE V2.0 dataset \cite{murakami2025adparaphrase}, which includes ad text pairs with human preference labels to study what makes ads more appealing. However, these resources are still limited to the text domain, and the field continues to lack broader and more comprehensive benchmarks that cover other types of marketing content and use cases. Future research is needed to develop standardized benchmarks and publicly available datasets that enable fair, consistent, and reproducible performance comparisons across different AI-generated marketing content methods.

\textbf{Culture-aware content generation.} Digital marketing content is often deployed across countries and regions where audiences have diverse cultural backgrounds. However, most current AI-generated digital marketing content methods are designed to optimize for a single objective, such as improving click-through rate, without explicitly modeling cultural context. As a result, content that performs well in one market may fail to engage users or even trigger negative reactions in other markets. For example, in data-driven and CTR-focused systems, certain elements may be selected because they consistently achieve high engagement in historical data from one cultural context. However, when applied to audiences from different cultural backgrounds, these same elements can carry unintended or even negative connotations, reducing their effectiveness or causing backlash. Therefore, there is a need for culture-aware content generation methods that can optimize for engagement metrics such as CTR while ensuring the produced content is culturally appropriate. Thus, an important direction for future work is to develop AI-generated marketing content methods that incorporate cultural context, allowing models to adjust their communication in ways that resonate with different audiences while maintaining strong engagement.

\subsection{Concluding statement}
AI-generated content is increasingly shaping the future of digital marketing by improving content creation efficiency and delivering better results through data-driven methods. Current research has made significant progress in text-based and image-based content generation, achieving improvements in metrics such as CTR. However, the field still faces key challenges, including the lack of unified methods for multi-modal content generation and the limited integration of user-specific preferences. Furthermore, comprehensive open benchmarks and datasets, as well as culture-aware content generation, remain important yet underexplored areas.

\section{Impact of AI-generated Content to Public Health 
}\label{sec:publichealth} 

Public healthcare is a constantly evolving domain that impacts the population at large. AI-generated content can play a critical role in assisting laypeople by providing essential information for addressing simple health concerns~\cite{branda2025role}. AI-based systems can also summarize extensive amounts of healthcare-related information into concise, accessible summaries that are easier for non-experts to understand~\cite{xie2025spreading}. In addition, such systems can deliver detailed information on specific healthcare topics, including medications, symptoms of various diseases, and related issues, thereby offering preliminary guidance to patients seeking knowledge in these areas~\cite{naumova2023responsible}. With internet access, ordinary people can interact with chatbot-style interfaces powered by AI, which can effectively assist them in managing basic to moderately complex health issues.

Access to healthcare remains a challenge for significant sections of the population in many countries, due to factors such as high costs and limited availability of qualified healthcare professionals~\cite{ozdemir2025people}. AI-generated content in public health has the potential to mitigate some of these barriers by offering accurate guidance and treatment suggestions for simple to mid-level conditions where no immediate risk of fatality exists. Such systems can help save both time and costs for individuals who face difficulties in accessing qualified medical practitioners.

Nevertheless, AI-based systems face considerable challenges, including the generation of hallucinated or inconsistent information, which can result in misleading or harmful outcomes for patients. Healthcare professionals also remain cautious about the direct use of AI in public healthcare, primarily because of its tendency to confidently produce incorrect or unreliable content. Detecting such inconsistencies is often difficult for both professionals and ordinary users~\cite{lehtimaki2024navigating}. Consequently, it is crucial to incorporate robust safety mechanisms before these systems are deployed at scale in public healthcare. Measures such as expert validation and cross-checking outputs across multiple AI systems are necessary to ensure that the disseminated information is reliable. Ultimately, the priority must be to develop trustworthy AI systems capable of addressing public healthcare challenges safely and effectively.



\subsection{Overview of Existing Studies}

Several studies~\cite{goktas2025shaping} have combined the expertise of medical professionals with the capabilities of AI tools to support clinical decision-making. Findings from these studies show mixed levels of trust and utility: while some clinicians place confidence in AI-generated content, others remain skeptical.
Other research~\cite{pantic2025artificial} has focused on training machine learning models, such as random forests, to detect AI-generated content in the domain of public health. In these studies, both AI-generated and human-authored paragraphs were compiled into datasets used for model training. Textual features were extracted using TF-IDF representations and subsequently employed to train the models.

Ethical concerns have also been raised regarding AI-generated content in public healthcare~\cite{naumova2023responsible}. A key issue relates to the authorship of textual content and outputs generated by LLMs, which are pre-trained on extensive public health data. Since this pre-training data often incorporates findings and publications from healthcare professionals and researchers, it raises important questions about ownership and attribution of AI-generated results.

Exploratory studies have further examined the contribution of AI to public health communication, including the ethical implications of crediting AI systems as co-authors in research papers.
In addition, research~\cite{branda2025role} on outbreaks of infectious diseases highlights the critical role of LLM-based chatbots in delivering timely and reliable information to the general public. These systems have been particularly effective in guiding individuals on protective measures during such crises.

\subsection{Research Challenges}
\textbf{Evidence fidelity and epidemiological correctness.}
One of the major research challenges lies in ensuring evidence fidelity and epidemiological correctness~\cite{augenstein2024factuality}. LLMs such as ChatGPT often produce fluent yet incorrect advice or casual, unsupported claims~\cite{sallam2023chatgpt}. Fact-checking the outputs of AI-based tools is difficult, as LLMs are prone to generating hallucinations -- small but significant pieces of fabricated or inconsistent content interwoven with correct information. Such outputs, while sounding credible, require careful supervision before acceptance. This challenge is particularly critical for public health, where fluently delivered but inaccurate information can have harmful consequences.

\textbf{Bias, equity, and cultural fitness.}
Medical content generated by LLMs is susceptible to biases, especially against low-resource languages, marginalized genders, castes, and regions~\cite{aneja2025beyond}. In contrast, these systems tend to privilege high-resource languages and urban contexts. The training data for most LLMs is heavily skewed toward English, urban, and high-resource environments~\cite{tripathi2025large}. As a result, individuals in rural areas, particularly those speaking local dialects, often receive less accurate or less contextually relevant health information. Furthermore, many regional languages are underrepresented in training corpora, leading models like ChatGPT to overlook structural realities such as limited access to medical facilities in rural settings. Instead, the models may generate advice based on assumptions of advanced healthcare infrastructure, which can be misleading for underserved populations.

\textbf{Health literacy alignment and accessibility.}
Another pressing concern is the mismatch between medical terminology used by LLMs and the health literacy levels of the general public. Since LLMs are often trained on data containing technical terms such as ``myocardial infarction'' rather than the more widely understood ``heart attack''\cite{esmaeilzadeh2020use}, their outputs may be inaccessible to laypeople\cite{yu2023zero}. Even when advice is medically sound, it may not be actionable in practice. For instance, recommendations like ``monitor your blood pressure regularly'' are unrealistic for households lacking access to a blood pressure machine. Similarly, instructions such as ``consult a cardiologist immediately'' may not be feasible in regions without specialist healthcare providers. This highlights the importance of aligning AI-generated advice with both the literacy and infrastructural realities of the target population.

\textbf{Ethical transparency and role clarity.}
Ethical transparency requires that AI-generated medical advice clearly state its sources, limitations, and scope~\cite{mohamed2024navigating}. Users need to know whether recommendations are based on international guidelines, local health policies, or probabilistic reasoning from generalized datasets. Without such transparency, public trust is at risk, especially if outputs later prove to be inaccurate or incompatible with local health practices. Transparency also extends to data privacy: users must be informed about how their data is collected, stored, and protected.

A further challenge concerns role clarity. AI-based models~\cite{esmaeilzadeh2020use} may produce medical advice that patients interpret as authoritative, despite the lack of clarity about the intended role of the system. If an AI model explicitly emphasizes its limitations, patients may dismiss it entirely; yet if it projects authority without disclaimers, they may follow potentially unsafe guidance. The key challenge is striking the right balance between providing useful, actionable advice and maintaining clear disclaimers about the system’s limitations.

\subsection{Research Propositions}

\textbf{Ensuring evidence fidelity and epidemiological correctness.}
Evidence fidelity and epidemiological correctness remain critical challenges in the use of AI-generated content for public health. AI systems that integrate structured fact-checking mechanisms, such as retrieval from validated medical sources, are more likely to demonstrate higher epidemiological correctness compared to unconstrained LLM outputs~\cite{agunlejika2025ai}. Similarly, the inclusion of citations and provenance indicators within AI-generated content enhances user trust and perceived reliability, even when the technical complexity of the information remains unchanged~\cite{agunlejika2025ai}. Moreover, supervised review of AI outputs by medical professionals substantially reduces the risk of hallucinations and inconsistent claims in public health advice, thereby ensuring stronger alignment with established medical evidence and epidemiological standards~\cite{branda2025role}. The central research question here is: \emph{how can transparent, trustworthy, and factually correct medical information be reliably generated by LLMs?}

\textbf{Balanced and context-Aware AI-generated medical content to mitigate bias and promote equity.}
Bias, equity, and cultural fitness are pressing concerns in the application of LLMs to medical content, as these systems often replicate and amplify inequities embedded in their training data~\cite{omar2025evaluating}. Since most LLMs are predominantly trained on English, urban, and high-resource contexts, they tend to produce outputs that disproportionately benefit those populations while underserving speakers of low-resource languages, rural communities, and marginalized genders, castes, or regions~\cite{rodriguez2024leveraging}. This imbalance creates significant challenges for individuals in rural areas who speak underrepresented dialects and lack access to advanced healthcare facilities, yet receive AI-generated advice that assumes such infrastructure exists~\cite{nazi2024large}. In contrast, AI systems fine-tuned on balanced datasets that include low-resource languages, rural dialects, and marginalized community contexts are more likely to produce culturally relevant and inclusive content. Additionally, incorporating explicit fairness constraints and bias-mitigation techniques into LLMs can help reduce disparities in the quality and relevance of medical advice across diverse populations. Embedding socio-contextual factors such as local healthcare infrastructure and cultural practices further enhances both equity and practical applicability of AI-generated outputs for underserved communities. The corresponding research question is: \emph{how can fairness constraints and bias-mitigation techniques be systematically incorporated into LLMs to reduce disparities in medical information across languages, genders, and regions?}

\textbf{Health literacy alignment and accessibility for common people.}
Improving health literacy alignment and accessibility is essential for ensuring the practical utility of AI-generated medical advice. Content that uses layperson-friendly language, adapts to local contexts, and provides resource-sensitive recommendations can significantly improve public understanding and real-world applicability, particularly in low-resource and underserved settings~\cite{okonji2024applications}. Moreover, medical information must account for the accessibility constraints of the patient seeking advice, including the availability of resources and facilities~\cite{george2024establishing}. The key research question here is: \emph{how can LLMs be designed to generate health-related information that is aligned with public health literacy and accessibility needs of the general population?}~\cite{goktas2025shaping}.

\textbf{Ethical transparency and role clarity.}
Ethical transparency requires that AI-generated medical systems clearly disclose their sources of knowledge (e.g., international guidelines, local health policies, or probabilistic reasoning from general datasets), articulate the scope and limitations of their outputs, and communicate data usage and privacy protections in a transparent manner~\cite{goktas2025shaping}. At the same time, achieving role clarity demands that systems strike a balance between projecting authority and providing appropriate disclaimers. Overstating authority risks patient overreliance, while excessive disclaimers may lead to disengagement. Carefully designed systems that manage this balance can foster informed decision-making, support patient autonomy, and ensure AI tools function as complements rather than replacements for professional medical advice. Such an approach can enhance trust, usability, and ethical alignment in diverse healthcare settings~\cite{george2024establishing}. The guiding research question is: \emph{how can LLMs clearly articulate the scope and limitations of their recommendations, transparently communicate data usage and privacy protections, and provide appropriate disclaimers without undermining user trust?}


\subsection{Concluding statement}
AI-generated content in the public health domain is a rapidly evolving phenomenon, with new techniques, research studies, and regulatory frameworks continually being developed to improve its usability for the general public. LLMs possess vast amounts of medical and general knowledge and have significant potential to address public health challenges, provided that the information they generate is user-friendly, trustworthy, factual, unbiased, aligned with health literacy, and ethically transparent. To realize this potential, researchers must design methods that ensure LLM-generated content is free from bias, incorporates ethical transparency and role clarity, aligns with health literacy needs, remains accessible to diverse populations, and reliably conveys accurate evidence and epidemiological information.

\section{Impact of AI-generated Content to Organizational Behaviour
}\label{sec:organizationalbehaviour} 
In traditional marketing teams, collaboration often depends on a clear division of work between strategists, analysts, and creatives. Organizational structures usually follow hierarchical models in which supervisors assign tasks and monitor outcomes. Such arrangements build trust by clarifying roles and responsibilities, but they can also create silos and slow communication. When consumer preferences change quickly, these rigid structures make it difficult for teams to cooperate across functions and to respond in a timely way.

With the rise of large language models and AI-generated content, AI is becoming an active participant in marketing practice. Collaboration shifts from rigid divisions of labour to more integrated workflows where humans and AI work together to solve tasks. At the same time, the growing presence of AI raises new questions about how trust is maintained and how organizational structures adapt to human–AI teaming. The following sections will review existing studies, highlight key challenges, and introduce research propositions on AI-generated content in organizational behaviour.

\subsection{Overview of Existing Studies}
The literature review identifies three key organizational behaviour themes in AI-generated marketing content, centering on collaboration, trust, and structural transformation. Effective human–AI teams require interoperability, trust-building, and mutual learning \cite{haupt2025consumer}, with augmented intelligence frameworks enhancing rather than replacing human capabilities \cite{french2024artificial}. This evolution reflects a shift from traditional supervisory models to collaborative teaming approaches that emphasize shared control and responsibility \cite{tsamados2024human}. However, realizing this vision demands cultural change and cross-functional cooperation among analysts, strategists, and creatives \cite{fenwick2024critical}. Consumer trust hinges on transparency, efficiency, and ethical AI use \cite{o2025consumer}, though negative perceptions persist despite disclosure \cite{haupt2025consumer} and responses vary by advertisement appeal \cite{chen2024consumer}. AI adoption faces psychological barriers including anticipatory and annihilation anxieties \cite{frenkenberg2025s}, while skepticism toward AI content influences purchase intentions through advertisement attitudes \cite{lu2025positive}. Organizationally, AI integration reshapes organizational structures and facilitates marketing strategy development \cite{jain2024artificial}, though career perception effects remain mixed \cite{yue2023impact}. As a strategic imperative rather than technical upgrade \cite{bezuidenhout2023impact}, successful implementation requires leaders to articulate clear visions, encourage dialogue, and empower employees, as poor leadership breeds confusion and disengagement \cite{hossain2025digital}.

\subsection{Research Challenges}\label{sec:researchchallenges_organ}
\textbf{Human Marketer–AI Collaboration.} A core challenge lies in designing human marketer-AI collaboration teams that maximize synergy through clear task allocation, interoperability, and mutual learning. While AI offers a path to enhancing human capabilities, its success depends on a significant cultural shift within marketing organizations. Encouraging collaboration across analysts, strategists, and creatives remains difficult, often hindered by skill gaps, unclear role boundaries, and resistance to change. Future research should explore practical interventions that strengthen AI fluency, promote cross-functional trust, and ensure productive human oversight.

\textbf{Trust and Ethics.} Sustaining consumer trust in AI-driven marketing is complicated by persistent negative perceptions, even when companies disclose AI use. Psychological barriers - such as fears of disruption or concerns over human identity - can undermine receptivity to AI-generated content. Research is needed to identify communication strategies and ethical frameworks that effectively reduce skepticism, tailor message appeal, and strengthen positive attitudes toward AI-mediated advertising. Cross-cultural and longitudinal studies could reveal how trust evolves over time and across diverse market contexts.

\textbf{Organizational Structure and Leadership.} The structural and leadership implications of AI adoption remain poorly understood. While AI can streamline processes and enable more adaptive marketing strategies, its impact on employee career perceptions and organizational design is mixed. Key challenges include aligning AI initiatives with broader strategic goals, maintaining employee engagement, and preventing confusion during transitions. Future research should investigate leadership practices, governance mechanisms, and change management approaches that facilitate smooth integration while safeguarding workforce morale. 

\subsection{Research Propositions}
Building on the organizational behaviour themes identified in Section \ref{sec:researchchallenges_organ}, the following 9 research propositions translate these challenges into hypotheses that address gaps in current research.

\textbf{Human Marketer–AI Collaboration.} 
(1) Marketing teams using hybrid content creation frameworks—where AI generates draft assets and humans iteratively refine messaging—achieve higher engagement and conversion rates than teams using solely human-created or fully AI-generated content. 
(2) AI platforms that provide transparent rationales for selecting specific creative elements in AI-delivered campaigns foster greater marketer acceptance and reduce pushback against AI-generated content recommendations. 
(3) The adoption of AI-delivered marketing tools shifts skill demand toward creative–analytical hybrid roles, where marketers interpret AI-driven consumer insights and shape narrative strategies around them. 
(4) Training programs that integrate AI literacy with hands-on AI content co-creation exercises increase marketers’ confidence and efficiency in producing high-quality, on-brand AI-generated marketing assets.

\textbf{Trust and Ethics.} 
(1) Transparent disclosure of AI training data provenance will increase consumer trust more effectively than generic statements about ethical AI use. 
(2) Consumers’ willingness to engage with AI-generated marketing content is enhanced when such content includes embedded “human endorsement cues” (e.g., CEO’s signatures or testimonials). 
(3) Personalized AI disclosure formats—tailored to individual psychological profiles—will outperform standardized disclosure messages in reducing skepticism toward AI-mediated advertising.

\textbf{Organizational Structure and Leadership.} 
(1) Marketing organizations adopting dual-leadership models—where a human leader and an AI system jointly oversee strategy—will experience higher innovation output than those led solely by humans or AI. 
(2) The clarity of AI’s decision-making boundaries within organizational structures moderates the relationship between AI adoption and employee engagement.

\subsection{Concluding statement}
While AI-generated content offers significant potential to enhance creativity, efficiency, and strategic adaptability in marketing, these benefits can only be fully realized through cultural transformation, transparent communication, and leadership vision. Addressing organizational behaviour barriers, clarifying role boundaries, and fostering trust - both internally and with consumers - are essential for sustaining long-term value. Ultimately, AI-generated content in marketing should be viewed not merely as a technical enhancement but as a strategic shift requiring coordinated human expertise, ethical stewardship, and structural alignment.

\section{Impact of AI-generated Content to Education
}\label{sec:education} 
As a maturing field, the education pathways into online marketing careers have also evolved. Higher education remains a central pathway for entry to the profession, particularly for roles that require strategic, analytical, or managerial capabilities. Undergraduate programs within business and commerce degrees increasingly embed digital marketing components, while postgraduate coursework programs often offer specialisations or standalone qualifications in business analytics, digital marketing, or data-driven marketing. 

As the workforce demands greater capabilities in digital platforms, performance metrics, and customer engagement strategies, higher education institutions have adapted their education programs to meet evolving industry needs \cite{parker2024}, including embedding emerging generative AI fluency \cite{grewal2025future}. However, the use of generative AI by students in higher education is producing broad concerns relating academic integrity, and the balance between new skill development and ensuring graduates attain key domain and problem-solving skills needed for successful careers.


In the following subsections, we examine prominent recent studies, highlight key challenges in online marketing education, and outline a research agenda to guide future developments of the field.  

\subsection{Overview of Existing Studies}

The increased integration of AI-generated content into online marketing practice presents both opportunities and challenges for marketing education. Recent research from Grewer, et al. \cite{grewal2025future} highlights the need to align marketing education with evolving industry practices, particularly in the development of digital automation and content generation using AI. Educational alignment is achieved through curriculum development, impacting the content of the material taught, as well as the mechanisms used to deliver this content. For example, Richter et al. \cite{Richter2025} emphasize the importance of integrating AI tools like ChatGPT and DALL·E into digital marketing curricula, arguing that active engagement with these tools bridges theoretical knowledge and practical application, preparing students for real-world marketing environments. Similarly, McAlister, Alhabash, and Yang \cite{mcalister2024artificial} explore the role of AI, and specifically ChatGPT, in marketing communications education, pointing to benefits such as enhanced learning efficiency, increased accessibility, and improved student engagement. 

The educational benefits of AI generated content is achieved through different use cases or roles. In terms of specific educational roles, Narang, Sachdev, and Liu \cite{Narang2025} propose that generative AI  can serve as a tutor, teammate, and tool within marketing education. As a tutor, AI may provide personalized feedback or guidance on concepts. As a teammate, it supports creative processes such as brainstorming, whereas as a tool, it can assist with data analysis and drafting or refining writing. This is a useful framing for considering how AI features in existing digital marketing education research.

Framed as a tutor, current research work positions generative AI as “knowledgeable” and “up-to-date” tutors that give explanations, examples and just-in-time guidance across marketing topics, with high reliability on widely known knowledge and more caution required on novel material \cite{ding2024generative}. In practice, Acar’s \cite{acar2024commentary} PAIR model operationalises AI-tutoring through a structured cycle (Problem, AI–Interaction, Reflection) that is human-centred, skill-centred and responsibility-centred, and pairs AI use with instructor monitoring so feedback and guidance remain scaffolded and equitable. Educator surveys and guidance similarly describe using generative AI to help with assignments, generating suggestions and providing feedback, reflecting growing acceptance of AI’s tutoring role in classrooms \cite{guha2024generative}. Complementing this, programmatic overviews emphasise “generative AI literacy", where students must learn to evaluate outputs, detect bias and understand limits, so AI tutoring augments rather than substitutes for human judgement \cite{schlegelmilch2025artificial}.

As a teammate, GenAI functions as a rapid ideation and co-creation partner that students collaborate with to generate and refine options. Instructors report that generative AI’s strength lies is the speed and volume of viable starting points for brainstorming, with better results when students prepare prompts and critique outputs. Specifically, marketing educators describe using generative AI alongside students to co-draft reports and marketing materials, collect background information, and iterate early concepts, explicitly mapping classroom teamwork with AI to how marketers collaborate with these tools in industry \cite{guha2024generative}.  Task taxonomies for the marketing classroom further position generative AI tools as collaborative partners across content generation, simulation and data analysis, enabling joint work on campaigns, plans and analytics that students can iteratively improve \cite{ding2024generative}.

Lastly, as a tool, AI is deployed to execute discrete education marketing and learning tasks at speed and scale. This involves tasks such as information provision and extraction, content editing and generation, simulation or role-play scenario script generation, and data analysis \cite{ding2024generative}.  In practice, generative AI tools are used to summarise market reports and literature, extract sentiment, automate routine course administration via custom assistants, and generate campaign artefacts such as copy and images, with staff and students verifying outputs \cite{ding2024generative} \cite{acar2024commentary}. Programmatic guidance recommends embedding these tools in project-based outputs (e.g., simulated data marketing plans, full campaigns) while addressing access and equity, and building generative AI literacy so students can evaluate, debug, and ethically use AI generated outputs \cite{guha2024generative} \cite{schlegelmilch2025artificial}. Finally,  research indicates that AI can generate useful analytics feedback in marketing coursework, producing quality rubrics and learning model choices, strengthening its role as an assessment and learning tool \cite{guha2024generative}.

\subsection{Research Challenges}
The research challenges that online marketing education is facing with the advent of AI are not very different from those in other fields of education. The researchers in all the work cited above have unanimously agreed and stressed the need for pedagogical adaptation in marketing education to keep pace with the rapid technological advancements. The current marketing curriculum lacks emphasis on AI literacy, including the skills and knowledge necessary for thriving in a technology-driven society \cite{Richter2025}. The gap between academic theory and industry practice is widening rapidly as AI is being integrated in novel ways in real-world applications at an unprecedented rate. 

A prominent theme in recent studies is that the integration of AI tools can radically enhance the marketing learning experience when used thoughtfully. As noted in the previous section, AI can take several roles in a marketing classroom, such as a personalised tutor, a teammate or resource tool, enhancing learners' autonomy, empowerment and enjoyment \cite{Narang2025}.   AI tools can also assist educators in diverse tasks such as engaging content creation, in-depth student data analysis, and developing personalised and adaptive curriculum, thereby significantly contributing to equity and accessibility in education. 

Despite its promises of transforming education, AI raises several pedagogical concerns for educators, and marketing academics are no exception. Academic integrity remains a central concern as AI challenges our current practice of assessing and evaluating learning outcomes \cite{mcalister2024artificial}. The focus must extend beyond merely ensuring the authenticity of student work; educators must also re-imagine what should be assessed and how assessment practices should evolve in an AI-driven educational landscape. Among other concerns are that the overreliance on AI may contribute to the potential erosion of foundational skills like critical thinking, problem solving, and creativity, which are essential for complex marketing tasks \cite{Narang2025}. To ensure effective and responsible integration of AI in marketing education, professional development and training of educators in AI pedagogy are essential \cite{Richter2025}.

The other significant challenge is embedding ethical considerations in all aspects of marketing education. Researchers emphasise the utmost importance of teaching students how to use AI tools responsibly in both educational and professional settings. Students must understand and effectively address the concerns related to data privacy, algorithmic bias, and misinformation, in a transparent and trustworthy manner in full compliance with legal, regulatory and ethical requirements \cite{grewal2025future,Narang2025}.  They must be taught to critically evaluate AI-generated outputs and learn to navigate safely in the complex ethical landscape surrounding the use of AI in digital marketing \cite{Richter2025,schlegelmilch2025artificial}.






\subsection{Research Propositions}
To remain relevant, curricula must become dynamic and agile, ensuring that graduates are prepared for the evolving job market and can adequately meet industry expectations. Marketing programs must complement the classical theories and foundational methodologies with AI-driven experiences throughout the learning process \cite{grewal2025future,guha2024generative}. AI should not be a standalone topic or course \cite{xia2023course}, but rather be integrated throughout the marketing curriculum. Essential new skills such as prompt engineering, AI-assisted data analysis, management of artificial influencers, and generative-AI tool-aided content creation, should be developed through marketing education, enabling graduates to apply AI effectively in real-world marketing scenarios \cite{schlegelmilch2025artificial}.

Marketing degree programs require significant overhauling, but there is also a crucial need for upskilling of the current marketing professionals who are grappling with rapid AI advancements. Universities must address this need by offering targeted short courses and flexible online training programs tailored for busy marketing executives and business leaders.  Through the lens of seeing generative AI as a tutor, team mate, or tool gives rise to the challenges associated with integrating AI in the digital marketing education landscape from these different perspectives, with each requiring distinctly different pedagogical strategies. 

The delivery of the curriculum by integrating new AI tools and AI-assisted analysis will necessitate pedagogical adaptation, significant effort and dedicated resources. There is an urgent need for developing novel conceptual frameworks for effective integration of generative AI in marketing education \cite{Narang2025}. The academics themselves will require targeted training and upskilling for the transformation of the curriculum and delivery of the redesigned curriculum. Therefore, the educational institutions must recognise this imperative and proactively commit to strategic investment for faculty and infrastructure development. 

Although several applications of AI are being tested for teaching marketing, our knowledge of the effect of such integration is limited. So a detailed agenda to explore the AI's roles and impact in marketing education is necessary, which should answer important research questions like, \textit{how does AI tutoring affect the teaching-learning experience, how does AI influence group dynamics as a team member, how does AI support and enhance personal learning experience, how can AI facilitate constructivist and experiential learning approaches}, etc. Besides, adoption studies are necessary to elucidate the factors influencing both educators' and learners' decisions and to develop strategies for mitigating the adoption barriers. Because of the rapidly evolving environment, continuous and close monitoring is necessary to ensure a deep and comprehensive understanding from both academics' and students' perspectives. 

Another line of research is necessary for developing robust frameworks to guide marketing professionals in using AI responsibly and ethically in real-world applications. Such frameworks must be developed with due consideration to the broader societal impacts of AI. Academia, industry and regulatory bodies should work together to ensure the relevance and practicality of such frameworks. Once established, such frameworks should be embedded in marketing programs to provide students with a solid foundation in the ethical and responsible use of AI in professional practice. 
%
%
%
%
\subsection{Concluding statement}
The rapid advancement of AI is driving a profound transformation in online marketing education, offering numerous opportunities and innovative solutions to long-standing challenges, both pedagogical and administrative. However, alongside these prospects, these technologies introduce significant challenges that are fundamentally reshaping how we teach and learn, assess educational outcomes, safeguard privacy, and uphold ethical and legal standards. Educators and institutions need to navigate this complex landscape carefully to harness AI's potential while safeguarding the integrity and inclusivity of marketing education. 

\section{Data Sovereignty and Security Risks in AI-generated Content
} \label{sec:datasovereignty}


The rise of AI-generated content(AIGC) in online marketing depends on vast datasets for training and operation, raising challenges in data sovereignty and security\cite{golda2024privacy, manduchi2024challenges}. Data sovereignty becomes exponentially complex when AIGC ingests, remixes and redistributes data streams. AIGC often utilize customer data and cultural references sourced across different areas, triggering conflicts between data fluidity and stringent regulatory frameworks such as GPDR, CCPA and emerging AI-specific legislation.

Meanwhile, the security vulnerabilities inherent in AIGC introduce attacks threatening data integrity and privacy. Security risks in generative AI not only contain raw data breaches, but also include model inversion attacks(reconstructing training data from outputs), prompt injection (manipulating AIGC to disclose sensitive information) and leakage of personal information through synthesized content. Online marketing applications amplify these threats by combining user data with generative models. As AIGC is embedded in customer-facing platforms, ensuring security becomes increasingly important.

\subsection{Overview of Existing Studies}

Existing research on data sovereignty and security risks in generative AI can be categorized into three primary paradigms: privacy-preserving technologies, adversarial defense mechanisms, and data sovereignty enforcement techniques. These approaches address distinct facets of the risk but exhibit significant trade-offs in practical deployment. 

\textbf{Privacy-preserving technologies}\cite{feretzakis2024privacy,liu2024generative,zhu2020more} constitute a primary approach to address data sovereignty and security risks in generative AI. Differential privacy(DP)\cite{wu2020privacy} injects noise into training data or gradients to mathematically bound privacy leakage, yet it degrades output quality. Federated learning(FL)\cite{sattler2019robust} distributes model training across local devices. However, statistical heterogeneity across clients impairs model convergence, while communication cost scales prohibitively with model size. Multi-party computation(MPC)\cite{zhang2021privacy} and trusted execution environments(TEEs) offer cryptographic isolation, which MPC via secret shared computations and TEEs via hardware enclaves. These techniques prevent direct data access but introduce significant latency.

\textbf{Adversarial defense mechanisms}\cite{chaudhury2021adversarial,vasan2020mthael}, containing adversarial training, input filtering and the development of robust model architectures, have been used to counteract adversarial attacks on generative models. Adversarial training\cite{zhao2022adversarial,bandi2023power} exposes generative models to adversarial samples to enhance their robustness, but reduces baseline accuracy. Input filtering\cite{ohm2024focusing,shivaani2025review} detects and filters out potentially malicious inputs before they reach the generative models. Robust architectural designs, such as self-attention masking, aim to intrinsically resist manipulation but hard to scale to multimodal systems where cross-modal vulnerabilities persist.

\textbf{Data sovereignty enforcement techniques}\cite{li2025data,ray2025assimilation} target provenance and access control. Digital watermarking\cite{lancaster2023watermark,takale2024watermark} embeds traceable markers such as encrypted origin signatures and content fingerprinting into AI-generated outputs. However, existing methods face scalability challenges in video and multimodal content. Attribute-based access control\cite{ragothaman2025access} restricts model access based on user roles and data sensitivity, yet requires centralized policy that conflicts with decentralized AI workflows.

\subsection{Research Challenges}

Resolving data sovereignty and security risks in AIGC faces persistent technical challenges, which can be categorized into the following three points.

\textbf{Sovereignty-functionality trade-off.} The intrinsic conflict between sovereignty requirements and model functionality remains a primary obstacle. Data sovereignty demands such as data localization of GDPR will lead to fragmented, jurisdiction-specific data, which directly conflicts with the global data integration essential for training high performance generative models. Although federated learning \cite{yuan2024robust,zheng2024decentralized,zheng2024poisoning,qu2024towards,yuan2025ptf} attempts to address this problem by keeping data local, it struggles with statistical heterogeneity across regions. 

\textbf{Training data memorization and reconstruction.} Recent works \cite{van2021memorization} show that publicly available LLMs and large-scale text-to-image models can implicitly ``memorize" training data. Model inversion attacks exploit this memorization tendencies, with studies confirming that samples from training dataset can be almost exactly reconstructed\cite{carlini2023extracting, somepalli2023diffusion, nasr2023scalable}. Although privacy-preserving technologies such as differential privacy offer attractive theoretical frameworks to ensure privacy, they still suffer from output quality deteriorates.

\textbf{Scalability of privacy-preserving techniques.} Existing privacy-preserving technologies fail to adapt to multimodal generative architectures. For example, in the context of image generation, scaling such approaches to highresolution remains elusive. Meanwhile, cryptography based solutions impose high latency for real-time applications such as personalized ad rendering, undermining user experience.

\subsection{Research Propositions}
\textbf{Integration with privacy constraints.}
Embedding privacy constraints in large-scale training of generative models can be a promising direction for further research. Future architecture could incorporate privacy-aware loss functions during pre-training that penalize memorization of sensitive information. Such constrained optimization may simultaneously minimize reconstruction risks demonstrated in model inversion attacks. For marketing applications such as personalized health campaigns, this could manifest as selective feature suppression, which automatically obscures protected attributes in representations while preserving contextual relevance.

\textbf{Integration with blockchain technology.}
Generative AI offers unparalleled capabilities in data processing and content creation, while blockchain provides the necessary framework to ensure the responsible and transparent use of these capabilities. For example, blockchain can improve data validity by cross-referencing multiple verified sources. AI-generated content can be compared with validated data to confirm its accuracy and verifiability. In addition, immutable terms can be set by combining smart contract with generative AI. Meanwhile, the transparency of blockchain can help to examine generative AI systems, which is essential in biases identification, accuracy and ethical verification.

\textbf{Integration with adversarial robustness.}
Integrating adversarial robustness into generative AI architectures shows the potential for securing content provenance and mitigating attacks. For example, information bottleneck framework minimizes redundant information while enhancing feature consistency between original and perturbed samples. Latent space watermarking can embed cryptographic signatures during generation to maintain traceability. Future work must consider lightweight robustness and standardized cross-modal defense to bridge the gap between generative capabilities and security demands. 

\subsection{Concluding statement}
The rise of AIGC in online marketing exposes data sovereignty requirements and security vulnerabilities. While some methods such as privacy-preserving technologies and adversarial defense mechanisms offer partial solutions, fundamental trade-offs persist in functionality and security. Future frameworks must integrate sovereignty compliance and attack resilience as core architectural principles to enable secure AIGC deployment without compromising functionality. 

\section{Conclusion}\label{sec:conclusion}
AIGC has demonstrated significant practical value by enhancing efficiency in content creation, improving information delivery, and driving innovative applications across domains such as digital marketing, education, and public health. Despite these rapid developments, there remain few studies that provide a systematic and cross-domain review of the latest progress and emerging challenges of AIGC. This vision paper aims to bridge this gap by bringing together insights from multiple disciplines to provide a cross-domain perspective on AIGC. It offers an overview of its technical foundations, spanning training, detection, and the spread and use of AI-generated content, and reviews its societal impacts across diverse domains. It further discusses unresolved challenges such as multimodal content generation, issues of bias and fairness of AIGC, and proposes research propositions to guide future work, including trust and ethical considerations as well as context-aware approaches. Moreover, it examines critical issues of data sovereignty and the security risks associated with generative AI.
Overall, this work offers a cross-domain perspective on AIGC that underscores its current trends, unresolved challenges, and the directions for future research.

\section{Acknowledgment}
This work is supported by the Australian Research Council under the streams of Discovery Project No. DP240101591.


\bibliographystyle{elsarticle-num} 
\bibliography{Sec1_IntroductionOverview,Sec3_ai-content-detection,Sec4_ai-content-spread-use,Sec5_ai-content-public-trust,Sec6_ai-content-digital-marketing,Sec7_ai-content-public-health,Sec8_ai-content-org-behavior,Sec9_ai-content-education,Sec10_data-sovereignty-security}







\end{document}